\documentclass{ecai}
\usepackage{times}
\usepackage{graphicx}
\usepackage{latexsym}
\usepackage{helvet} 
\usepackage{courier}  
\usepackage[hyphens]{url}  

\usepackage{amsmath}
\usepackage{amsfonts}
\usepackage{amstext}
\usepackage{booktabs}
\usepackage{algorithm}
\usepackage{multirow}
\usepackage{color}
\usepackage{framed}
\usepackage{array}
\usepackage[table,xcdraw]{xcolor}
\usepackage{booktabs}
\usepackage[noend]{algpseudocode}
\usepackage{algorithmicx}
\usepackage{amssymb}

\ecaisubmission   

\begin{document}

\title{Joint Extraction of Entities and Relations Based on \\ a Novel Decomposition Strategy}

\author{
Bowen Yu$^{1,2}$
\and Zhenyu Zhang$^{1,2}$
\and Xiaobo Shu$^{1,2}$
\and Tingwen Liu$^{1*}$ 
\\ Yubin Wang\thanks{Institute of Information Engineering, Chinese Academy of Sciences, Bejing, China. Emails: \{yubowen, zhangzhenyu1996, shuxiaobo, liutingwen, wangyubin\}@iie.ac.cn. * Corresponding Author.} $^{\,,}$\thanks{School of Cyber Security, University of Chinese Academy of Sciences, Bejing, China.}
\and Bin Wang\thanks{Xiaomi AI Lab, Xiaomi Inc., China. Email: wangbin11@xiaomi.com.}
\and Sujian Li\thanks{Key Laboratory of Computational Linguistics, Peking University, MOE, China. Email: lisujian@pku.edu.cn.}}

\maketitle
\bibliographystyle{ecai}

\begin{abstract}
Joint extraction of entities and relations aims to detect entity pairs along with their relations using a single model.
Prior work typically solves this task in the extract-then-classify or unified labeling manner.
However, these methods either suffer from the redundant entity pairs, or ignore the important inner structure in the process of extracting entities and relations.
To address these limitations, in this paper, we first decompose the joint extraction task into two interrelated subtasks, namely HE extraction and TER extraction.
The former subtask is to distinguish all head-entities that may be involved with target relations, and the latter is to identify corresponding tail-entities and relations for each extracted head-entity. 
Next, these two subtasks are further deconstructed into several sequence labeling problems based on our proposed span-based tagging scheme, which are conveniently solved by a hierarchical boundary tagger and a multi-span decoding algorithm.
Owing to the reasonable decomposition strategy, our model can fully capture the semantic interdependency between different steps, as well as reduce noise from irrelevant entity pairs.
Experimental results show that our method outperforms previous work by 5.2\%, 5.9\% and 21.5\% (F1 score), achieving a new state-of-the-art on three public datasets.
\end{abstract}

\section{INTRODUCTION}
Extracting pairs of entities with relations from unstructured text is an essential step in automatic knowledge base construction, and an ideal extraction system should be be capable of extracting overlapping relations (i.e., multiple relations share a common entity) \cite{zeng2018extracting}.
Traditional pipelined approaches first recognize entities, then choose a relation for every possible pair of extracted entities.
Such framework makes the task easy to conduct, but ignoring the underlying interactions between these two subtasks \cite{li2014incremental}.
One improved way is to train them jointly by parameter sharing \cite{fu-etal-2019-graphrel,miwa2016end,sun2019joint}.
Although showing promising results, these extract-then-classify approaches still require explicit separate components for entity extraction and relation classification.
As a result, \emph{their relation classifiers may be misled by the redundant entity pairs} \cite{dai2019joint,tan2019jointly}, since \emph{N} entities will lead to roughly \emph{N}$^2$ pairs, and most of which are in the NA (non-relation) class.

\begin{figure*}
\centerline{\includegraphics[width=0.90\textwidth]{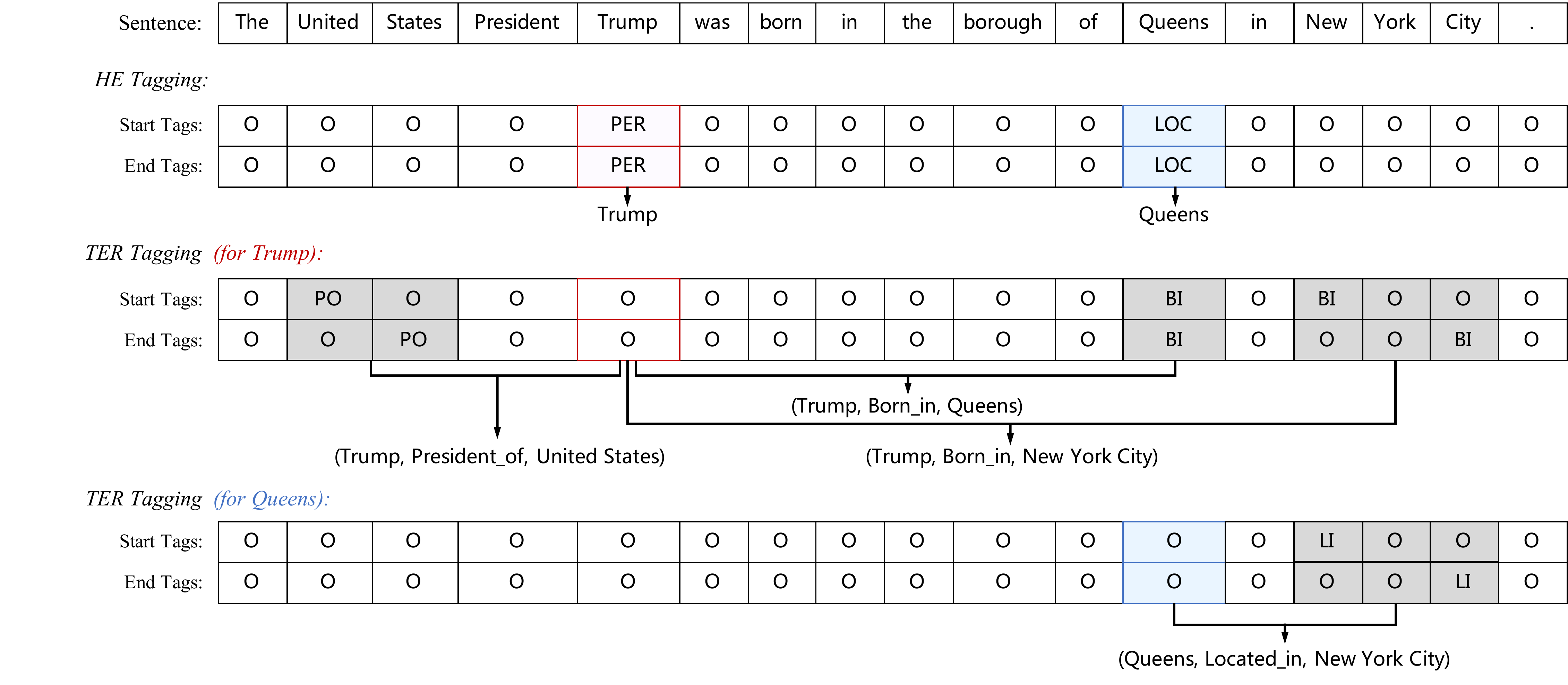}}
    \caption{An example of our tagging scheme. \emph{PER} is short for entity type \emph{PERSON}, \emph{LOC} is short for \emph{LOCATION}, \emph{PO} is short for relation type \emph{President\_of}, \emph{BI} is short for \emph{Born\_in}, and \emph{LI} is short for \emph{Located\_in}.}\label{fig:tagging}
\end{figure*}

Rather than extracting entities and relations separately, Zheng et al. \cite{zheng2017joint} proposed a unified tagging scheme to transform joint extraction to a sequence labeling problem with a kind of multi-part tags.
However, this model lacks the elegance to identify overlapping relations, which may lead to poor recall when processing a sentence with overlapping relations.
As the improvement, Dai et al. \cite{dai2019joint} presented PA-LSTM which tags entity and relation labels simultaneously according to each query word position, and achieves state-of-the-art performance.
Nevertheless, \emph{these models always ignore the inner structure such as dependency included in the head entity, tail entity and relation due to the unified labeling-once process}.
As is well known, a tail-entity and a relation should be depended on a specific head-entity. 
In other words, if one model cannot fully perceive the semantics of head-entity, it will be unreliable to extract the corresponding tail entities and relations.

For a complex NLP task, it is very common to decompose the task into easier modules or processes, and a reasonable design is quite crucial to help one model make further progress \cite{hu2019open,liu2018empower,zhang2019sentiment}.
In this paper, through analysis of the two kinds of methods above, we exploit the inner structure of joint extraction and propose a novel decomposition strategy, which hierarchically decomposes the task into several sequence labeling problems with partial labels capturing different aspects of the final task (see Figure \ref{fig:tagging}).
Starting with a sentence, we first judiciously distinguish all the candidate head-entities that may be involved with target relations, then label corresponding tail-entities and relations for each extracted head-entity.
We call the former subtask as \textbf{H}ead-\textbf{E}ntity (HE) extraction, and the later as \textbf{T}ail-\textbf{E}ntity and \textbf{R}elation (TER) extraction.
Such extract-then-label (ETL) paradigm can be understood by decomposing the joint probability of triplet extraction into conditional probability $p(h,r,t|S) = p(h|S)p(r,t|h,S)$, where $(h,r,t)$ is a triplet in sentence $S$. 
In this manner, \emph{our TER extractor is able to take the semantic and position information of the given head-entity into account when tagging tail-entities and relations}, and naturally, one head-entity can interact with multiple tail-entities to form overlapping relations. 
Besides, compared with the extract-then-classify methods, \emph{our paradigm no longer extracts all entities at the first step, only head-entities that are likely to participate in target triplets are identified, thus alleviating the impact of redundant entity pairs}.

Next, inspired by extractive question answering which identifies answer span by predicting its start and end indices \cite{seo2016bidirectional}, we further decompose HE and TER extraction with a span-based tagging scheme. 
Specifically, for HE extraction, entity type is labeled at the the start and end positions of each head-entity.
For TER extraction, we annotate the relation types at the start and end positions of all the tail-entities which have relationship to a given head-entity. 
To enhance the association between boundary positions, we present a hierarchical boundary tagger, which labels the start and end positions separately in a cascade structure and decode them together by a multi-span decoding algorithm.
By this means, HE and TER extraction can be modeled in the unified span-based extraction framework, differentiated only by their prior knowledge and output label set.
Overall, for a sentence with $m$ head-entities, the entire task is deconstructed into $2+2m$ sequence labeling subtasks, the first $2$ for HE tagging and the other $2m$ for TER.
Intuitively, the individual subtasks are significantly easy to learn compared with the whole extraction task, suggesting that by trained cooperatively with shared underlying representations, they can constrain the learning problem and achieve a better overall outcome.

We evaluate our method on three public datasets: NYT-single, NYT-multi and WebNLG. Experimental results show that the proposed method significantly outperforms previous work on normal, overlapping and multiple relation extraction, increasing the SOTA F1 score to 59.0\% (+5.2\%), 78.0\% (+5.9\%) and 83.1\% (+21.5\%), respectively.
Further analysis confirms the effectiveness and rationality of our decomposition strategy.

\section{METHODOLOGY}
\label{sec:method}

In this section, we first introduce our tagging scheme, based on which the joint extraction task is transformed into several sequence labeling problems. 
Then we detail the hierarchical boundary tagger, which is the basic labeling module in our method. 
Finally, we move on to the entire extraction system.

\subsection{Tagging Scheme}
\label{sec:scheme}

Let us consider the head-entity (HE) extraction first. As discussed in the previous section, it is decomposed into two sequence labeling subtasks.
The first subtask mainly focuses on identifying the start position of one head-entity. One token is labeled as the corresponding entity type if it is the start word, otherwise it is assigned the label ``\emph{O}'' (Outside).
In contrast, the second subtask aims to identify the end position of one head-entity and has a similar labeling process except that the entity type is labeled for the token which is the end word.

For each identified head-entity, TER extraction is also decomposed into two sequence labeling subtasks which make use span boundaries to extract tail-entities and predict relations simultaneously.
The first sequence labeling subtask mainly labels the relation type for the token which is the start word of the tail-entity, while the second subtask tags the end word.

Figure \ref{fig:tagging} illustrates an example of our tagging scheme, in which the words ``\emph{United}'', ``\emph{States}'', ``\emph{Trump}'', ``\emph{Queens}'', ``\emph{New}'' and  ``\emph{City}'' are all related to final extraction results, thus they are labelled with special tags. 
For example, the word ``\emph{Trump}'' is the first and also the last word of head-entity ``\emph{Trump}'', so the tags are both \emph{PERSON} in the start and end tag sequences when tagging HE. 
For TER extraction, when the given head-entity is ``\emph{Trump}'', there are two tail-entities involved in with wanted relations, i.e., \emph{(``Trump'', President\_Of, ``United States'')} and \emph{(``Trump'', Born\_In, ``New York City'')}, so ``\emph{United}'' and ``\emph{New}'' are labeled as \emph{President\_Of} and \emph{Born\_In} respectively in the start tag sequences. Similarly, we can obtain end tag sequences that ``\emph{States}'' and ``\emph{City}'' are marked. 
Beyond that, the other words irrelevant to the final result are labeled as ``\emph{O}''.

Note that our tagging scheme is quite different from PA-LSTM \cite{dai2019joint}. 
For an $n$-word sentence, PA-LSTM builds $n$ different tag sequences according to different query position while our model tags the same sentence for $2+2\times m$ times to recognize all overlapping relations, where $m$ is the number of head-entities and $m<<n$. 
This means our model is more time-saving and efficient. 
Besides, it uses ``BIES''  signs to indicate the position of tokens in the entity while we only predict the start and end positions without loss of the ability to extract multi-word entity mentions.

\subsection{Hierarchical Boundary Tagger}

\begin{figure*}
  \centering
  \includegraphics[width=0.95\textwidth]{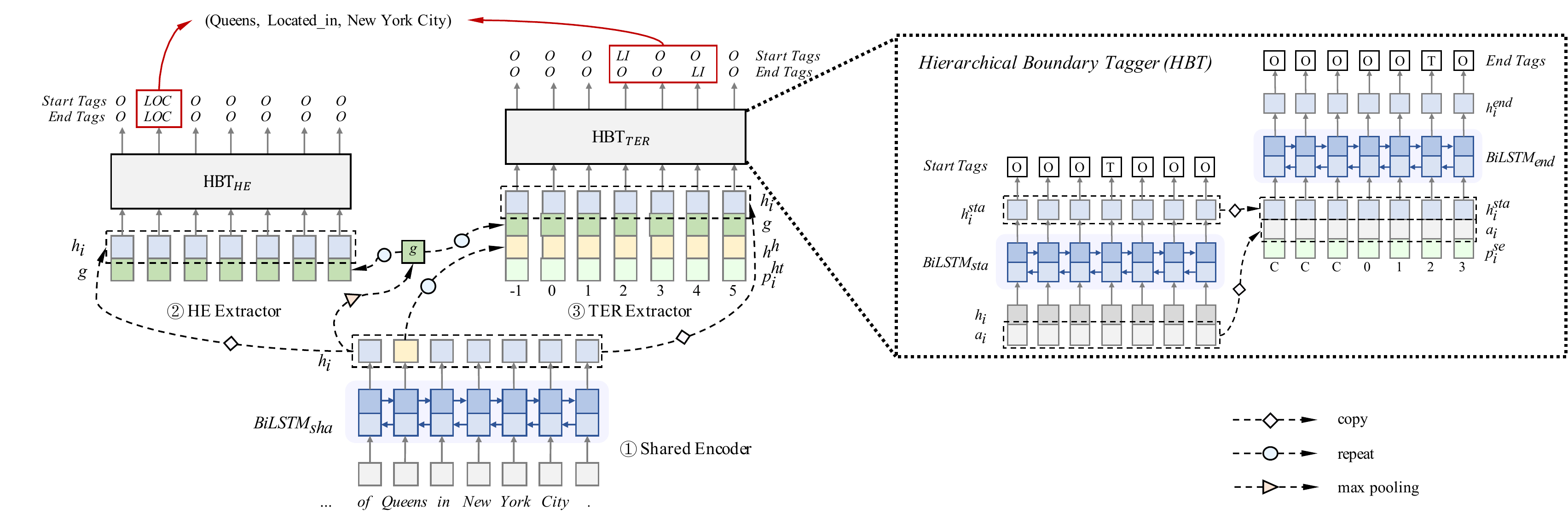}
    {\caption{An illustration of our model. The left panel is an overview of our joint extraction system, and the right panel shows the detailed structure of our sequence tagger HBT. Here, ``\textit{Queens}'' is extracted by the HE extractor, then its hidden state in the shared encoder is marked as the yellow box and entered into the TER  extractor as prior knowledge.
    }
    \label{fig:system}}
\end{figure*}

According to our tagging scheme, we utilize a unified architecture to extract HE and TER. 
In this paper, we wrap such extractor into a general module named hierarchical boundary tagger (abbreviated as HBT). 
For the sake of generality, we do not distinguish between head and tail-entity, and they are collectively referred to as targets in this subsection. 
Formally, the probability of extracting a target $t$ with label $l$ (entity type for head-entity or relation type for tail-entity) from sentence $S$ is universally modeled as:
\begin{equation}
\label{equ:motiva2}
   p(t,l|S) = p(s_{t}^{l}|S)p(e_{t}^{l}|s_{t}^{l},S)
\end{equation}
where $s_{t}^{l}$ is the start index of $t$ with label $l$ and $e_{t}^{l}$ is the end index. 
Such decomposition indicates that there is a natural order among the tasks: predicting end positions may benefit from the prediction results of start positions, which motivates us to employ a hierarchical tagging structure.
As shown in the right panel of Figure \ref{fig:system}, we associate each layer with one task and take the tagging results as well as hidden states from the low-level task as input to the high-level. 
In this work, we choose BiLSTM \cite{hochreiter1997long} as the basic encoder.
Formally, the label of word $x_i$ when tagging the start position is predicted as Eq. \ref{equ:sta-tag}.

\begin{equation}
\label{equ:sta-input}
   \boldsymbol{\rm{h}}^{sta}_i = {\rm{BiLSTM}}_{sta}([\boldsymbol{\rm{h}}_i;\boldsymbol{\rm{a}}_i])
\end{equation}
\begin{equation}
P(y^{sta}_i) = {\rm{Softmax}}(\boldsymbol{\rm{W}}^{sta}\cdot \boldsymbol{\rm{h}}^{sta}_i + \boldsymbol{\rm{b}}^{sta})
\end{equation}
\begin{equation}
\label{equ:sta-tag}
\text{sta\_tag}(x_i)=\mathop{\arg\max}_{k} P(y^{sta}_i=k)
\end{equation}

\noindent where $\boldsymbol{\rm{h}}_i$ denotes token representation and $\boldsymbol{\rm{a}}_i$ is an auxiliary vector. 
For HE extraction, $\boldsymbol{\rm{a}}_i$ is a global representation learned from the entire sentence. 
It is beneficial to make more accurate predictions from a global perspective. 
For TER extraction, $\boldsymbol{\rm{a}}_i$ is the concatenation of a global representation and a head-entity-related vector to indicate the position and semantic information of the given head-entity.  
Here we adopt ${\rm{BiLSTM}}_{sta}$ to fuse $\boldsymbol{\rm{h}}_i$ with $\boldsymbol{\rm{a}}_i$ into a single vector $\boldsymbol{\rm{h}}^{sta}_i$. 
Analogously, $x_i$'s end tag can be calculated by Eq. \ref{equ:prob-end-tag}.
\begin{equation}
\label{equ:end-input}
   \boldsymbol{\rm{h}}^{end}_i = {\rm{BiLSTM}}_{end}([\boldsymbol{\rm{h}}^{sta}_i;\boldsymbol{\rm{a}}_i;\boldsymbol{\rm{p}}_i^{se}])
\end{equation}
\begin{equation}
\label{equ:prob-end-tag}
P(y^{end}_i) = {\rm{Softmax}}(\boldsymbol{\rm{W}}^{end}\cdot \boldsymbol{\rm{h}}^{end}_i + \boldsymbol{\rm{b}}^{end})
\end{equation}
\begin{equation}
\label{equ:end-tag}
\text{end\_tag}(x_i)=\mathop{\arg\max}_{k} P(y^{end}_i=k)
\end{equation}

The difference between Eq. \ref{equ:sta-input}-\ref{equ:sta-tag} and Eq. \ref{equ:end-input}-\ref{equ:end-tag} is twofold. 
Firstly, we replace  $\boldsymbol{\rm{h}}_i$ in Eq. \ref{equ:sta-input} with  $\boldsymbol{\rm{h}}^{sta}_i$ to make model aware of the hidden states of start positions when predicting end positions. 
Secondly, inspired by the position encoding vectors used in \cite{zeng2014relation}, we feed the position embedding $\boldsymbol{\rm{p}}_i^{se}$ to the ${\rm{BiLSTM}}_{end}$ layer as its additional input. 
$\boldsymbol{\rm{p}}_i^{se}$ can be obtained by looking up $p^{se}_i$ in a trainable position embedding matrix, where 
\begin{equation}
\label{equ:pe-hbm}
p^{se}_i =
\begin{cases}
i-s^{*}, & \mbox{if } s^* \ \mbox{exists}\\
C, & \mbox{otherwise}
\end{cases} 
\end{equation}

Here $s^{*}$ is the nearest start position before current index, and $p^{se}_i$ is the relative distance between $x_i$ and $s^{*}$. 
When there is no start position before $x_i$, $s^{*}$ will not exist, then $p^{s}_i$ is assigned as a constant $C$ that is normally set to the maximum sentence length. 
In this way, we explicitly limit the length of the extracted entity and teach model that the end position is impossible to be in front of the start position. 
To prevent error propagation, we use the gold $p^{se}$ (distance to the correct nearest start position) during training process.

We define the training loss (to be minimized) of HBT as the sum of the negative log probabilities of the true start and end tags by the predicted distributions:
\begin{equation}
\label{equ:hbm}
\small
   \mathcal{L}_{\text{\emph{HBT}}} = -\frac{1}{n} \sum_{i=1}^n (\log P(y^{sta}_i=\hat{y}^{sta}_i)+\log P(y^{end}_i=\hat{y}^{end}_i))
\end{equation}

\noindent where $\hat{y}^{sta}_i$ and $\hat{y}^{end}_i$ are the true start and end tags of the $i$-th word, respectively, and $n$ is the length of the input sentence.

\begin{algorithm}[t]
\label{algo:hbm}
\small
\caption{Multi-span decoding} 
\label{algo:hmsd}
{\bf Input:} \hspace*{0.02in} \\ 
$S$, $C$ \\
\hspace*{0.15in} $S$ denotes the input sentence\\
\hspace*{0.15in} $C$ is a predefined distance constant\\
{\bf Output:} \hspace*{0.02in}\\ 
$\{(e_j,tag_j)\}_{j=1}^m$, \\
\hspace*{0.15in} $e_j$ denotes the $j$-th extracted target and $tag_j$ is the type tag
\begin{algorithmic}[1]
\State Define $n \leftarrow$ Sentence Length%
\State Initialize $\mathbf{R} \leftarrow \{\}$%
\State Initialize $s^{*} \leftarrow 0$ %
\State Initialize $p^{se}$ as a  list of length $n$ with default value $C$%
\State Obtain \text{sta\_tag}$(S)$ by Eq. \ref{equ:sta-tag} %
\For{$idx \leftarrow 1$ to $n$} %
  \If{sta\_tag $(S)[idx] \neq $ ``\emph{O}''}
      \State $s^{*} \leftarrow idx$ 
  \EndIf
  \If{$s^{*}>0$}
    \State $p^{se}[idx] \leftarrow idx-s^{*}$ 
  \EndIf
\EndFor 
\State Obtain $\boldsymbol{\rm{p}}^{se}$ by transforming $p^{se}$ into matrix %
\State Obtain \text{end\_tag}$(S)$ by Eq. \ref{equ:end-tag}%
\For{$idx_{s}\leftarrow 1$ to $n$} %
  \If{\text{sta\_tag}$(S)[idx_{s}] \neq $ ``\emph{O}''}
    \For{$idx_{e}\leftarrow idx_{s}$ to $n$ } %
      \If{\text{end\_tag}$(S)[idx_{e}] = $ \text{sta\_tag}$(S)[idx_{s}]$}
      \State $e \leftarrow S[idx_{s}:idx_{e}]$
      \State $tag \leftarrow $\text{end\_tag}$(S)[idx_{e}]$
      \State $\mathbf{R} \leftarrow \mathbf{R} \cup \{(e,tag)\}$
      \State Break
      \EndIf
    \EndFor 
  \EndIf
\EndFor 
\State \Return $\mathbf{R}$
\end{algorithmic}
\end{algorithm}

At inference time, to adapt to the multi-target extraction task, we propose a multi-span decoding algorithm, as shown in Algorithm 1. 
For each input sentence $S$, we first initialize several variables (Lines 1-4) to assist with the decoding: (1) $n$ is defined as the length of $S$. (2) $\mathbf{R}$ is initialized as an empty set to record extracted targets and type tags. (3) $s^{*}$ is introduced to hold the nearest start position before current index. (4) $p^{se}$ is initialized as a list of length $n$ with default value $C$ to save the position sequence $[p^{se}_1,\cdots,p^{se}_n]$. 
Next, we obtain the start tag sequence by Eq. \ref{equ:sta-tag} (Line 5) and compute $p^{se}_i$ for each token by Eq. \ref{equ:pe-hbm} (Lines 6-10). 
On the basis of $p^{se}$, we can get $\boldsymbol{\rm{p}}^{se}$ by looking up position embedding matrix (Line 11) . 
Then the tag sequence of end position can be computed by Eq. \ref{equ:end-tag} (Line 12). 

Now, all preparations necessary are in place, we start to decoding \text{sta\_tag}$(S)$ and \text{end\_tag}$(S)$. 
We first traverse \text{sta\_tag}$(S)$ to find the start position of a target (Line 13). 
If the tag of current index is not ``\emph{O}'', it denotes that this position may be a start word (Line 14), then we will traverse \text{end\_tag}$(S)$ from this index to search for a end position (Line 15). 
The matching criterion is that if the tag of the end position is identical to the start position (Line 16), the words between the two indices are considered to be a candidate target (Line 17), and the label of start position (or end position) is deemed as the tag of this target (Line 18). 
The extracted target along with its tag is then added to the set $\mathbf{R}$ (Line 19), and the search in \text{end\_tag}$(S)$ is terminated to continue to traverse \text{sta\_tag}$(S)$ to find the next start position (Line 20). 
Once all the indices in \text{sta\_tag}$(S)$ are iterated, this decoding function ends by returning the recordset $\mathbf{R}$ (Line 21).

\subsection{EXTRACTION SYSTEM}

With the span-based tagging scheme and the hierarchical boundary tagger, we propose an end-to-end neural architecture (Figure \ref{fig:system}) to extract entities and overlapping relations jointly, which first encodes the sentence with a shared BiLSTM encoder. 
Then, a HE extractor is built to extract head entities. 
For each extracted head entity, the TER extractor is triggered with this head-entity's semantic and position information to detect corresponding tail-entities and relations.

\subsubsection{Shared Encoder} 
Given sentence $S=\{x_1,\cdots,x_n\}$, we utilize a BiLSTM layer to incorporate information from both forward and backward directions:
\begin{equation}
\label{equ:base-lstm}
   \boldsymbol{\rm{h}}_i = {\rm{BiLSTM}}_{sha}(\boldsymbol{\rm{x}}_i)
\end{equation}
where $\boldsymbol{\rm{h}}_i$ is the hidden state at position $i$, and $\boldsymbol{\rm{x}}_i$ is the word representation of $x_i$ which contains pre-trained embeddings and character-based word representations generated by running a CNN on the character sequence of $x_i$. 
 Following~\cite{fu-etal-2019-graphrel}, we also employ part-of-speech (POS) embedding to enrich $\boldsymbol{\rm{x}}_i$.

\subsubsection{HE Extractor}
HE extractor aims to distinguish candidate head-entities and exclude irrelevant ones. We first concatenate $\boldsymbol{\rm{h}}_i $ and $\boldsymbol{\rm{g}}$ to get the feature vector $\tilde{\boldsymbol{\rm{x}}}_i=[\boldsymbol{\rm{h}}_i;\boldsymbol{\rm{g}}]$, where $\boldsymbol{\rm{g}}$ is a global contextual representation computed by max pooling over all hidden states. 
Actually, $\boldsymbol{\rm{g}}$ works as the $\boldsymbol{\rm{a}}_i$ for each token in Eq. \ref{equ:sta-input}.
Moreover, we use $\boldsymbol{\rm{H}}_{\text{\emph{HE}}} = \{\tilde{\boldsymbol{\rm{x}}}_1, \cdots, \tilde{\boldsymbol{\rm{x}}}_n\}$ to denote all the word representations for HE extraction and subsequently feed $\boldsymbol{\rm{H}}_{\text{\emph{HE}}}$ into one HBT to extract head-entities:
\begin{equation}
   \mathbf{R}_{\text{\emph{HE}}} = {\rm{HBT}}_{\text{\emph{HE}}}(\boldsymbol{\rm{H}}_{\text{\emph{HE}}})
\end{equation} 
\noindent where  ${\mathbf{R}_{\text{\emph{HE}}}=\{(h_j,type_{h_j})\}_{j=1}^m}$  contains all the head-entities and corresponding entity type tags in $S$.

\subsubsection{TER Extractor} 

Similar to HE extractor, TER extractor also uses the basic representation $\boldsymbol{\rm{h}}_i$ and global vector $\boldsymbol{\rm{g}}$ as input features. 
However, simply concatenating $\boldsymbol{\rm{h}}_i$ and $\boldsymbol{\rm{g}}$ is not enough for detecting tail-entities and relations with the specific head-entity.
The key information required to perform TER extraction includes: 
(1) the words inside the tail-entity; 
(2) the depended head-entity; 
(3) the context that indicates the relationship; 
(4) the distance between tail-entity and head-entity. 
Under these considerations, we propose the position-aware, head-entity-aware and context-aware representation $\bar{\boldsymbol{\rm{x}}}_i$. 
Given a head-entity $h$, we define $\bar{\boldsymbol{\rm{x}}}_i$ as follows:
\begin{equation}
   \bar{\boldsymbol{\rm{x}}}_i = [\boldsymbol{\rm{h}}_i;\boldsymbol{\rm{g}};\boldsymbol{\rm{h}}^{h};\boldsymbol{\rm{p}}_i^{ht}]
\end{equation}
where $\boldsymbol{\rm{h}}^{h}=[\boldsymbol{\rm{h}}_{s_{h}};\boldsymbol{\rm{h}}_{e_{h}}]$ denotes the representation of head-entity $h$, in which $\boldsymbol{\rm{h}}_{s_{h}}$ and  $\boldsymbol{\rm{h}}_{e_{h}}$ are the hidden states at the start and end indices of $h$ respectively.
 $\boldsymbol{\rm{p}}_i^{ht}$ is the position embedding to encode the the relative distance from $x_i$ to $h$. 
Obviously,   $[\boldsymbol{\rm{g}};\boldsymbol{\rm{h}}^{h};\boldsymbol{\rm{p}}_i^{ht}]$ is the auxiliary feature vector for TER extraction as $\boldsymbol{\rm{a}}_i$ in Eq. \ref{equ:sta-input}.

Formally, we take $\boldsymbol{\rm{H}}_{\text{\emph{TER}}}$ $=$ $\{\bar{\boldsymbol{\rm{x}}}_1, \cdots, \bar{\boldsymbol{\rm{x}}}_n\}$ as input to one HBT, and the output $\mathbf{R}_{\text{\emph{TER}}}$ $=$ $\{(t_o,rel_{o})\}_{o=1}^z$, in which $t_o$ is the $o$-th extracted tail-entity and $rel_{o}$ is its relation tag with the given head-entity. 
\begin{equation}
   \mathbf{R}_{\text{\emph{TER}}} = {\rm{HBT}}_{\text{\emph{TER}}}(\boldsymbol{\rm{H}}_{\text{\emph{TER}}} )
\end{equation} 

Then we can assemble triplets by combining $h$ and each $(t_o,rel_{o})$ to form $\{(h,rel_{o},t_{o})\}_{o=1}^z$, which contains all triplets with head-entity $h$ in sentence $S$ \footnote{Note that $type_{h}$ is not included in the final output of our extraction system. However, we claim that by predicting entity types, we can implicitly incorporate type information into head-entity representation, which is beneficial to the subsequent TER tagging as our experiment reveals.}.
It is worth noting that at the training time, $h$ is the gold head-entity, while at the inference time we select head-entity one by one from $\mathbf{R}_{\text{\emph{HE}}}$ to complete the extraction task.

\subsubsection{Training of Joint Extractor} 
Two learning signals are provided to train the model: $\mathcal{L}_{\text{\emph{HE}}}$ for HE extraction and $\mathcal{L}_{\text{\emph{TER}}}$ for TER extraction, both are formulated as Eq.\ref{equ:hbm}.
To share input utterance across tasks and train them jointly, for each training instance, we randomly select one head-entity from gold head-entity set as the specified input of the TER extractor. 
We can also repeat each sentence many times to ensure all triplets are utilized, but the experimental results show that this is not beneficial. 
Finally, the joint loss is given by: 
\begin{equation}
\label{equ:loss}
\mathcal{L} = \mathcal{L}_{\text{\emph{HE}}} + \mathcal{L}_{\text{\emph{TER}}}
\end{equation} 

Then, the model is trained with stochastic gradient descent.
Optimizing Eq.\ref{equ:loss} enables the extraction of head-entity, tail-entity, and relation to be mutually influenced, such that, errors in each component can be constrained by the other.

\section{EXPERIMENTS}

\subsection{Experimental Settings}

\subsubsection{Datasets}

\begin{table}
\begin{center}
{\caption{Statistics of the datasets.}
    \label{datasets}}
    \begin{tabular}{lrrr}
    \toprule
        Dataset & NYT-single & NYT-multi & WebNLG \\
    \midrule
        \# Relation types & 24 & 24 & 246 \\
        \# Training sentences & 66,335 & 56,195 & 5,019 \\
        \# Test sentences & 395 & 5,000 & 703 \\
        
    \bottomrule
    \end{tabular}
\end{center}
\end{table}

Following popular choices and previous work~\cite{dai2019joint,fu-etal-2019-graphrel,ren2017cotype,zeng-etal-2019-learning,zeng2018extracting,zheng2017joint}, We conduct experiments on three benchmark datasets:
(1) \textbf{NYT-single} is sampled from the New York Times corpus \cite{riedel2010modeling} and published by Ren et al~\cite{ren2017cotype}. The training data is automatically labeled using distant supervision, while 395 sentences are annotated manually as test data, most of which have single triplet in each sentence. 
(2) \textbf{NYT-multi} is published by Zeng et al.~\cite{zeng2018extracting} for testing overlapping relation extraction, they selected  5000 sentences from NYT-single as the test set, 5000 sentences as the validation set and the rest 56195 sentences are used as training set. 
(3) \textbf{WebNLG} is proposed by Claire et al.~\cite{gardent2017creating} for Natural Language Generation task. We use the dataset pre-processed by Zeng et al~\cite{zeng2018extracting} and the train set contains 5019 sentences, the test set contains 703 sentences and the validation set contains 500 sentences. Statistics of the datasets are shown in Table \ref{datasets}. 

Besides, as suggested in \cite{fu-etal-2019-graphrel,zeng2018extracting}, we also divided the test set into three categories: Normal, SingleEntityOverlap (SEO), and EntityPairOverlap (EPO) to verify the effectiveness on extracting overlapping relations. 
Specifically, a sentence belongs to Normal class if none of its triplets has overlapping entities. 
If the entity pairs of two triplets are identical but the relations are different, the sentence will be added to the EPO set. 
A sentence belongs to SEO class if some of its triplets have an overlapped entity and these triplets don’t have any overlapped entity pair.
Note that a sentence in the EPO set may contain multiple Normal and SEO triplets.
We discuss the result for different categories in the detailed analysis.

\subsubsection{Evaluation}
We follow the evaluation metrics in previous work \cite{dai2019joint,fu-etal-2019-graphrel,ren2017cotype,zeng-etal-2019-learning,zeng2018extracting,zheng2017joint}.
A triplet is marked correct if and only if its relation type and two corresponding entities are all correct, where the entity is considered correct if the head and tail offsets are both correct. 
We adopt the standard micro Precision, Recall and F1 score to evaluate the results.

\subsubsection{Implementation Details}
We use the 300 dimension Glove \cite{pennington2014glove} to initialize word embeddings. 
The POS, character and position embeddings are randomly initialized with 30 dimensions. 
The window size of CNN for character-based word representations is set to 3, and the number of filters is 50. 
For Bi-LSTM encoder, the hidden vector length is set to 100. Parameter optimization is performed using Adam \cite{kingma2014adam} with learning rate 0.001 and batch size 64. 
Dropout is applied to word embeddings and hidden states with a rate of 0.4. 
To prevent the gradient explosion problem, we set gradient clip-norm as 5. 
All the hyper-parameters are tuned on the validation set. 
We run 5 times for each experiment then report the average results.

\begin{table*}
\begin{center}
{
    \caption{Main results on three benchmark datasets. Bold marks highest number among all models. $\ddagger$ marks results 
    quoted directly from the original papers. $^\dagger$ marks results reported in \cite{dai2019joint} and \cite{zeng2018extracting}. $^*$ marks results produced with offcial implementation. }
    \label{main}
}	
    \setlength{\tabcolsep}{2mm}{\begin{tabular}{lccccccccc}
    \toprule
        \multirow{2}{*}{Model} & \multicolumn{3}{c}{NYT-single 
               } & \multicolumn{3}{c}{NYT-multi}& \multicolumn{3}{c}{WebNLG}\\
         & Precision & Recall & F1 & Precision & Recall & F1 & Precision & Recall & F1 \\
    \midrule
        CoType$^\ddagger$ \cite{ren2017cotype} & 42.3\% & 51.1\% & 46.3\% & -- & -- & -- & -- & -- & -- \\
        NovelTagging$^\dagger$ \cite{zheng2017joint} & \textbf{61.5}\% & 41.4\% & 49.5\% & 32.8\% & 30.6\% & 31.7\% & 52.5\% & 19.3\% & 28.3\% \\
        MultiDecoder$^\ddagger$ \cite{zeng2018extracting} & -- & -- & -- & 61.0\% & 56.6\% & 58.7\% & 37.7\% & 36.4\% & 37.1\% \\
        MultiHead$^*$ \cite{bekoulis2018joint} & 51.5\% & 52.8\% & 52.1\% & 60.7\% & 58.6\% & 59.6\% & 57.5\% & 54.1\% & 55.7\% \\
        PA-LSTM$^\ddagger$ \cite{dai2019joint} & 49.4\% & 59.1\% & 53.8\% & -- & -- & -- & -- & -- & -- \\
        GraphRel$^\ddagger$ \cite{fu-etal-2019-graphrel} & -- & -- & -- & 63.9\% & 60.0\% & 61.9\% & 44.7\% & 41.1\% & 42.9\% \\
        OrderRL$^\ddagger$ \cite{zeng-etal-2019-learning} & -- & -- & -- & 77.9\% & 67.2\% & 72.1\% & 63.3\% & 59.9\% & 61.6\% \\
    \midrule
        ETL-BIES & 51.1\% & 64.6\% & 57.2\% & 84.4\% & 71.5\% & 77.4\% & 83.5\% & 81.1\% & 82.3\%\\
        ETL-Span & 53.8\% & \textbf{65.1\%} & \textbf{59.0\%} & \textbf{85.5\%} & \textbf{71.7\%} & \textbf{78.0\%} & \textbf{84.3} \% & \textbf{82.0\%} & \textbf{83.1\%}\\
    \bottomrule
    \end{tabular}}
\end{center}
\end{table*}

\subsubsection{Comparison Models}

For comparison, we employ the following models as baselines: 
(1) \textbf{Cotype} \cite{ren2017cotype} learns jointly the representations of entity mentions, relation mentions and type labels;
(2) \textbf{NovelTagging} \cite{zheng2017joint} is the first proposed unified sequence tagger which predicts both entity type and relation class for each word; 
(3) \textbf{MultiDecoder} \cite{zeng2018extracting} considers relation extraction as a sequence-to-sequence problem and uses dynamic decoders to extract relation triplets; 
(4) \textbf{MultiHead} \cite{bekoulis2018joint} first identifies all candidate entities, then perform relation extraction by identifying multiple relations for each entity, these two tasks are trained jointly;
(5) \textbf{PA-LSTM} \cite{dai2019joint} is the current best unified labeling method, which tags entity and relation labels simultaneously according to a query word position and achieves the recent state-of-the-art results on the NYT-single dataset; 
(6) \textbf{GraphRel} \cite{fu-etal-2019-graphrel} is the latest extrat-then-classify method, which first employs GCNs to extract hidden features, then predicts relations for all word pairs of an entity mention pair extracted by a sequence tagger;
(7) \textbf{OrderRL} \cite{zeng-etal-2019-learning} is the state-of-the-art method on the NYT-multi and WebNLG datasets, which applies the reinforcement learning into a sequence-to-sequence model to generate multiple triplets.

We call our proposed extract-then-label method with span-based scheme as \textbf{ETL-Span}. 
In addition, to access the performance influence of span-based scheme, we also implement another competitive baseline by replacing our tagger with widely used BiLSTM-CRF without any change in the input features ($\tilde{\boldsymbol{\rm{x}}}_i$ and $\bar{\boldsymbol{\rm{x}}}_i$), and utilize BIES-based scheme accordingly, which associates each type tag (entity type or relation type) with four position tags to indicate entity positions and types simultaneously, denoted as \textbf{ETL-BIES}.

\subsection{Experimental Results and Analyses}

\subsubsection{Main Results}

\begin{table}
\begin{center}
{\caption{ Comparison of test-time speed. Bat/s refers to the number of batches can be processed per second.}\label{tab:speed}}
    \setlength{\tabcolsep}{2mm}{\begin{tabular}{lccccccccc}
    \toprule
    
         Model & NYT-single & NYT-multi & WebNLG \\
         \midrule
        ETL-BIES & 10.9 Bat/s  & 11.2 Bat/s & 6.3 Bat/s \\
        ETL-Span & 26.1 Bat/s  & 25.6 Bat/s & 23.5 Bat/s \\
    \bottomrule
    \end{tabular}}
\end{center}
\end{table} 

\begin{table}
\begin{center}
{\caption{An ablation study of ETL-Span on the NYT-multi dev set.}
\label{tab:ablation}}
\begin{tabular}{lccc}
  \toprule
      Model                & Precision & Recall & F1 \\
  \midrule
  ETL-Span                & \textbf{86.5\%} & \textbf{73.5\%} & \textbf{79.5\%}    \\
  \quad-- Char embedding  & 83.1\% & 71.2\% & 76.7\%   \\
  \quad-- Position embedding $\boldsymbol{\rm{p}}^{ht}$  & 81.9\% & 70.3\% & 75.7\%    \\
  \quad-- Hierarchical tagging & 84.6\% & 70.7\% & 77.0\%    \\
  \quad-- Head-entity type tagging & 85.8\% & 72.2\% & 78.4\%    \\
    \quad-- Joint learning & 80.4\% & 68.9\% & 74.2\%    \\
  \bottomrule
\end{tabular}
\end{center}
\end{table}

Table \ref{main} reports the results of our models against other baseline methods. 
It can be seen that our method, ETL-Span, significantly outperforms all other methods and achieves the state-of-the-art F1 score on all three datasets.
Over the latest extract-then-classify method GraphRel, ETL-Span achieves substantial improvements of 16.1\% and 40.2\% in F1 score on the NYT-multi and WebNLG datasets respectively. 
We attribute the performance gain to two design choices:  
(1) the integration of tail-entity and relation extraction as it captures the interdependency between entity recognition and relation classification; 
(2) the exclusion of redundant (non-relation) entity pairs by the judicious recognition of head-entities which are likely to take part in some relations.
For the NYT-single dataset, ETL-Span improves by a relative margin of 5.2\% against the strong baseline PA-LSTM. 
We consider that it is because
(1) we decompose the difficult joint extraction task into several more manageable subtasks and handle them in a mutually enhancing way;
(2) our TER extractor effectively captures the semantic and position information of the depended head-entity, while PA-LSTM detects tail-entities and relations relying on a single query word.
In addition, we find that the results of our model are better than sequence-to-sequence methods like MultiDecoder and OrderRL, it is likely due to the innate restrictions on RNN unrolling, the capacity of generating triplets is limited \cite{fu-etal-2019-graphrel}.
Beyond that, we notice that the Precision of our model drops compared with NovelTagging on the NYT-single dataset. 
One possible reason is that many overlapping relations are not annotated in the manually labeled test data.
Following PA-LSTM \cite{dai2019joint}, we add some gold triplets into NYT-single test set and further achieve a large improvement of 12.5\% in F1 score and 18.7\% in Precision compared with the results in Table \ref{main}.
Overall, these results indicate that our extraction paradigm which first extracts head-entity then labels corresponding tail-entity and relation can better capture the relational information in the sentence.

We also observe that ETL-Span performs remarkably better than ETL-BIES, we guess it is because ETL-BIES must do additional work to learn the semantics of the BIES tags, while in ETL-Span, the entity position is naturally encoded by the set of type labels, thus reducing the tag space of each functional tagger.
Another advantage of span-based tagging is that it avoids the computing overhead of CRF, as shown in Table \ref{tab:speed}, ETL-Span accelerates the decoding speed of ETL-BIES by up to 3.7 times.
  The main reason is that decoding the best chain of labels with CRF requires a significant amount of computing resources especially when the tag space is huge (e.g., on WebNLG with 246 relations and 989 tags). 
Besides, ETL-Span only takes about 1/4 time per batch and 1/5 GPU memory compared with ETL-BIES during training, which further verdicts the superiority of our span-based scheme.

 \subsubsection{Ablation Study}

  To demonstrate the effectiveness of each component, we remove one particular component at a time to understand its impact on the performance. Concretely, we investigated character embedding, position embedding $\boldsymbol{\rm{p}}^{ht}$, hierarchical tagging (by tagging boundary positions at the outmost BiLSTM layer), head-entity type tagging (by tagging 0/1 instead of entity types in the HE extractor) and joint learning (by training HE extractor and TER extractor separately without parameter sharing).
  From these ablations shown in Table \ref{tab:ablation}, we find that: 
 (1) Consistent with PA-LSTM \cite{dai2019joint}, the character-level representations are helpful to capture the morphological information and deal with OOV words.
 (2) When we remove $\boldsymbol{\rm{p}}^{ht}$, the score drops by 3.8\%, which indicates that it is vital to let tail-entity extractor aware of position information of the given head-entity to filter out irrelevant entities by implicit distance constraint. 
  (3) Removing the hierarchical tagging structure hurts the result by 2.5\% F1 score, which indicates that predicting end positions benefits from the prediction results of start positions. 
  (4) By predicting entity type in the HE extractor, we can implicitly incorporate type information into head-entity representation, which is beneficial to the subsequent TER tagging. 
  (5) Compared with the pipelined manner, joint learning framework brings a remarkable improvement (5.3\%) in F1 score, which demonstrates that our HE extractor and TER extractor actually work in the mutual promotion way, and again confirms the effectiveness and rationality of our decomposition strategy.

\subsubsection{Analysis on Different Sentence Types }
\begin{figure}[t]
    \centering
    \includegraphics[width=0.7\linewidth]{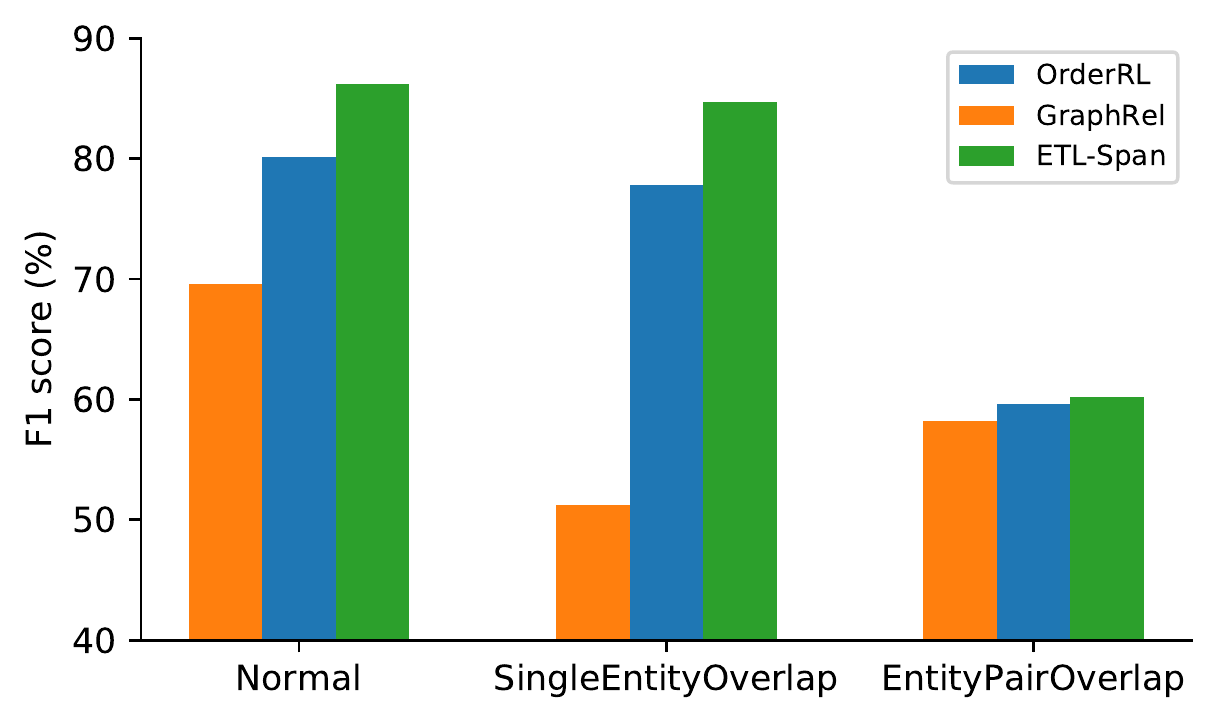}
    \caption{ F1 score by overlapping category on the NYT-multi test set.} 
     \label{fig:overlap}
\end{figure}

To verify the ability of our model in handling the overlapping problem, following \cite{fu-etal-2019-graphrel,zeng2018extracting}, we conduct further experiments on the NYT-multi test set.
The results are shown in Figure \ref{fig:overlap}.
Among the compared baselines, GraphRel and OrderRL are the latest two models with the capacity to handle the EPO triplets. For this purpose, GraphRel predicts relations for all word pairs, in this case, its relation classifier will be overwhelmed by the superfluous candidates. 
OrderRL utilizes a sequence-to-sequence model to decode overlapping relations but can decode only the first word of multi-word entity, while ours can detect the whole.
Readers may have noticed that ETL-Span cannot solve the problem of entity pair overlapping. 
Nevertheless, ETL-Span still surpasses baselines in all categories. 
Specifically, ETL-Span outperforms OrderRL by 6.1\% on the Normal class, 6.9\% on the SEO class, and 0.6\% on the EPO class.
In fact, even on the EPO set, there are still a significant amount of triplets where entity pairs don't overlap.
The most common triplets in the real-life corpus are those of Normal and SEO class and our substantial surpass on these two categories masks our shortcomings on the EPO class. We leave the identification of EPO triplets for future work.

\begin{figure}[t]
    \centering
    \includegraphics[width=0.7\linewidth]{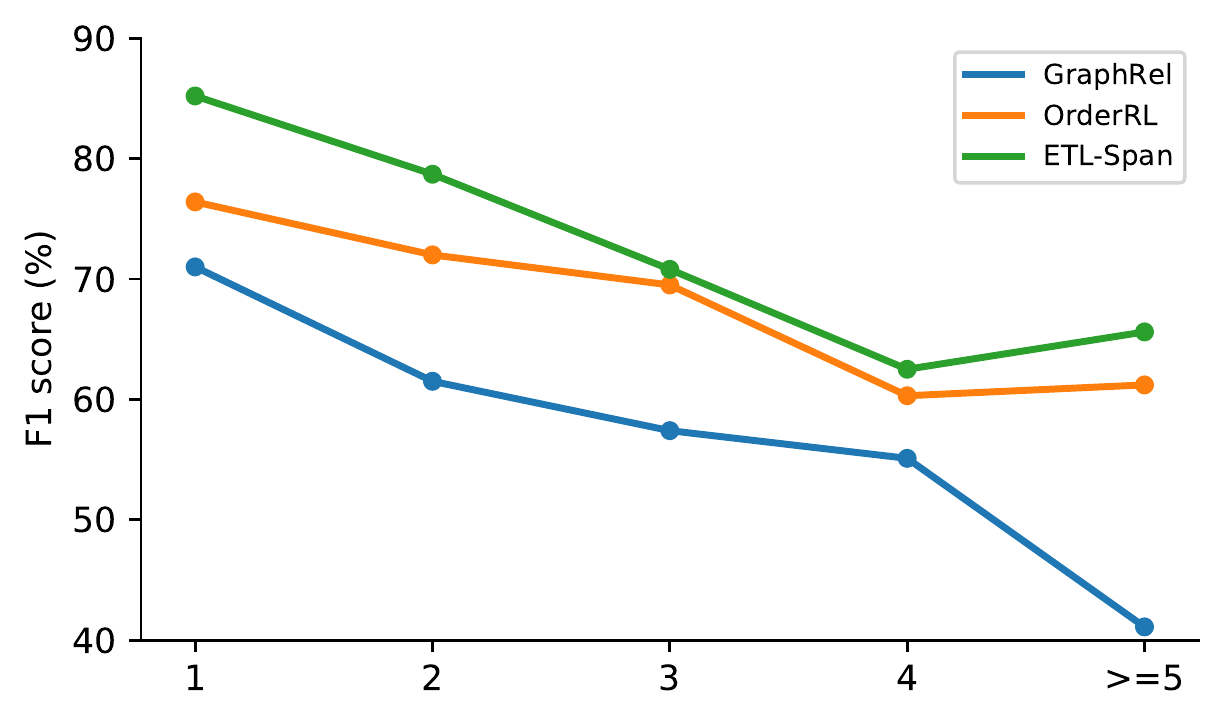}
    \caption{F1 score by sentence triplet count on the NYT-multi test set.} 
     \label{fig:tripcnt}
\end{figure}

We also compare the results given different numbers of triplets in a sentence, and sentences in the NYT-multi test set are divided into 5 subclasses, each class contains sentences that has 1,2,3,4 or $\geq$ 5 triplets. 
As illustrated in Figure \ref{fig:tripcnt}, ETL-Span outperforms the baselines under all numbers of triplets in a sentence.
When the sentence only contains one triplet, ETL-Span yields a 8.8\% improvement in comparison with OrderRL.
When there are multiple triplets in a sentence, ETL-Span still outperforms GraphRel and OrderRL significantly.
These observations demonstrate that our extraction paradigm is effective to handle the multiple relation extraction task.



\section{RELATED WORK}
There have been extensive studies for entity relation extraction task. 
Traditional pipelined methods divide this task into two separate subtasks: first extract the token spans in the text to detect entity mentions, and then discover the relational structures between entity mentions \cite{zelenko2003kernel}.
Entity recognization has traditionally been solved as a sequence labeling problem, and most recent work leverages a LSTM-CRF architecture \cite{lample2016neural,mengge2019porous,zhang2018chinese}. 
Relation extraction is normally treated as a problem of multi-label classification \cite{zhang2020distilling}. 
Zeng et al. \cite{zeng2014relation} employed a deep convolutional neural network for extracting lexical and sentence level features. 
Zhou et al. \cite{zhou2016attention} combined attention mechanisms with BiLSTM to reduce intra-sentence noise.
Yu et al. \cite{yu2019beyond} proposed to learn the latent relational expressions based on the segment attention layer for relation extraction.
However, all these methods require preprocessing step such as NER and ignore interactions between entity recognization and relation extraction, therefore may suffer from error propagation \cite{bekoulis2018joint, dai2019joint,ren2017cotype,sun2018extracting}.

To address the above limitation, a variety of joint learning methods were proposed \cite{gupta2016table,katiyar2017going,li2019entity,takanobu2019hierarchical,zhang2017end}. 
Kate et al. \cite{kate2010joint} presented a card-pyramid graph to represent entities and their relations in a sentence.
Miwa et al. \cite{miwa2014modeling} introduced a simple and flexible table representation of entities and relations.
However, these models need complicated process of feature engineering, which requires much manual efforts and domain expertise. 
Recently, several end-to-end neural architectures are applied to joint relation extraction.
Sun et al. \cite{sun2018extracting} optimized a global loss function to jointly train entity recognition model and relation classification model under the framework work of Minimum Risk Training.
Bekoulis et al. \cite{bekoulis2018adversarial,bekoulis2018joint} first recognized the entities, then they formulated the relation extraction task as a multi-head selection problem. For each entity, they calculated the score between it and every other entities for a given relation.
Tan et al. \cite{tan2019jointly} first identified all candidate entities, then performed relation extraction via ranking with translation mechanism.
Sun et al. \cite{sun2019joint} developed an entity-relation bipartite graph to perform joint inference on entity types and relation types. 
Fu et al. \cite{fu-etal-2019-graphrel} utilized graph convolutional network to extract overlapping relations by splitting entity mention pairs into several word pairs and considering all pairs for prediction.
Nevertheless, these extract-then-classify methods still require explicit separate components for entity extraction and relation classification, and the relation classifier may be overwhelmed by the redundant extracted entity pairs.
Another line of work \cite{zeng-etal-2019-learning,zeng2018extracting} directly generated triplets one by one by a sequence-to-sequence model but fail to extract an entity that has multiple words.
Zheng et al. \cite{zheng2017joint} proposed a unified tagging model which utilizes a special tagging scheme to convert joint extraction task to a sequence tagging problem. However, their model cannot recognize overlapping relations in the sentence.
As the improvement, Dai et al. \cite{dai2019joint} proposed to extract overlapping triplets by tagging one $n$-word sentence for $n$ times. 
Unfortunately, due to the labeling-once process, this kind of unified labeling methods cannot fully exploit the inter-dependency between entities and relations.

In this paper, we design a novel joint extraction paradigm which first extracts head-entities and then labels tail-entities and relations for each head-entity. In essence, it bridges the gap between extract-then-classify and unified labeling approaches.
More specifically, when compared with the extract-then-classify methods, our extract-then-label paradigm no longer extracts all entities at the first step, only head-entities that are likely to participate in target triplets are identified, thus alleviating the impact of redundant entity pairs.
Owing to the reasonable decomposition strategy, our model can better capture the correlations between head-entities and tail-entities than unified labeling approaches, thus resulting in a better joint extraction performance.
Besides, our span-based tagging scheme is inspired by recent advances in machine reading comprehension \cite{seo2016bidirectional}, which derived the answer by predicting its start and the end indices in the paragraph. 
Hu et al. \cite{hu2019open} also applied this sort of architecture to open-domain aspect extraction and achieved great success.

\section{CONCLUSIONS}

In this paper, we present an end-to-end sequence labeling framework for joint extraction of entities and relations based on a novel decomposition strategy.
Experimental results show that the functional decomposition of the original task simplifies the learning process and leads to a better overall learning outcome, achieving a new state-of-the-art on three public datasets.
Further analysis demonstrates the ability of our model in handling normal, overlapping and multiple relation extraction.
In the future, we would like to explore similar decomposition strategy in other information extraction tasks, such as event extraction and aspect extraction.
The source code of this paper can be obtained from https://github.com/yubowen-ph/JointER.


\section{ACKNOWLEDGEMENTS}
The authors thank Jianlin Su from Zhuiyi Tech Co. Ltd.for his helpful discussions. 
The work presented in this paper is supported by the National Key Research and Development Program of China (grant No.2016YFB0801003) and the Strategic Priority Research Program of Chinese Academy of Sciences (grant No.XDC02040400).

\bibliography{ecai}
\end{document}